\documentclass{bmvc2k}
\usepackage{amssymb}
\usepackage{booktabs}
\usepackage{twemojis}
\usepackage{wrapfig}
\usepackage{makecell}

\title{HQ-JEPA: Hybrid Quantum Joint-Embedding Predictive Architecture for Cross-Modal Remote Sensing Representation Learning}

\addauthor{Md Aminur Hossain$^{*}$}{md.aminurhossain@gmail.com}{1}
\addauthor{Ayush V. Patel$^{*}$}{ayu020503@gmail.com}{2}
\addauthor{Sanjay K. Singh}{sks@sac.isro.gov.in}{1}
\addauthor{Biplab Banerjee}{getbiplab@gmail.com}{2}

\addinstitution{
 Space Applications Centre,\\
 Indian Space Research Organisation, \\
 Ahmedabad, India
}
\addinstitution{
 Centre of Studies in Resources Engineering,\\
 Indian Institute of Technology Bombay,\\
 Mumbai, India
}

\runninghead{Hossain et al.}{HQ-JEPA: Hybrid Quantum JEPA for Remote Sensing}

\begin{document}

\maketitle
\footnotetext{$^{*}$ Equal Contribution.}

\begin{abstract}
We introduce HQ-JEPA, a hybrid quantum-classical joint-embedding predictive architecture for cross-modal remote sensing representation learning. The proposed framework extends JEPA-style masked latent prediction to paired Sentinel-1 and Sentinel-2 imagery by predicting masked target representations from visible context regions while aligning heterogeneous modality features in a shared embedding space. To improve representation quality, HQ-JEPA combines four complementary objectives: latent token prediction, cross-modal token alignment, SIGReg-based Gaussian regularization in the fused latent space, and a differentiable SWAP-test-based Fidelity Quantum Similarity (FQS) loss. Unlike pixel reconstruction methods, HQ-JEPA learns semantic representations directly in latent space and uses quantum state-overlap-based similarity as an additional regularization signal. We evaluate the pretrained encoder on GeoBench classification and segmentation tasks under linear probing and fine-tuning settings. Results show that HQ-JEPA achieves competitive and often superior performance over strong self-supervised and remote sensing foundation-model baselines, demonstrating the benefit of integrating predictive self-supervision, cross-modal geometric regularization, and quantum fidelity-based representation learning for remote sensing applications.
\end{abstract}

\section{Introduction}
\label{sec:intro}

Self-Supervised Learning (SSL)~\cite{gui2024survey} provides an efficient way to learn valuable representations from visual data without human annotations. By exploiting the intrinsic structure of images, SSL enables scalable pretraining and improves downstream performance in tasks such as classification~\cite{huang2023self, chen2021self}, object detection~\cite{huang2022survey}, and image segmentation~\cite{ouyang2022self}. Recent advances in masked image modeling~\cite{hondru2025masked} and Joint-Embedding Predictive Architectures (JEPAs)~\cite{brotee2025survey} have led to models that yield coherent, semantically meaningful features by predicting latent representations instead of raw pixel data. This encourages the model to learn semantically meaningful representations that are more transferable and robust.

Although SSL has achieved strong performance in unimodal settings, many real-world applications require learning from heterogeneous modalities~\cite{zong2024self}. This is particularly important in remote sensing, where optical and synthetic aperture radar (SAR) observations provide complementary information about the same scene. Aligning such modalities without manual supervision remains challenging due to differences in imaging physics, spectral response, spatial texture, and noise characteristics~\cite{farhadizadeh2025challenges}. Existing cross-modal SSL methods typically rely on contrastive objectives~\cite{jaiswal2020survey} or regression-based alignment losses~\cite{navaneet2022simreg}, which encourage instance-level agreement but do not explicitly regulate the geometry or distribution of the learned embedding space. Consequently, a large modality gap often leads to poorly conditioned, fragile, or collapsed representations~\cite{jing2021understanding, he2022exploring}.

\begin{wrapfigure}{r}{0.45\columnwidth}
    \vspace{-10pt}
    \centering
    \includegraphics[width=0.43\columnwidth]{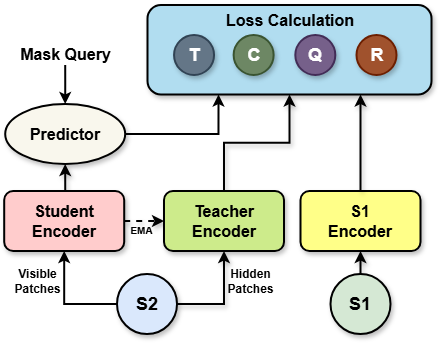}
    \caption{Overview of HQ-JEPA.}
    \label{fig:teaser}
    \vspace{-10pt}
\end{wrapfigure}

The role of representation geometry in the generalization and stability of machine learning models has been widely studied in recent years~\cite{cosentino2022geometry}. The results show that structured embedding distributions improve transfer performance and prevent collapse in non-contrastive self-supervised learning~\cite{sansone2024collapse}. However, the effects of distributional regularization on cross-modal predictive learning have received little attention. Moreover, the most commonly used approach to measure similarity between different modalities is the use of cosine and Euclidean distances, which mainly capture pairwise geometric agreement and may be insufficient to model nonlinear relationships between heterogeneous modality representations~\cite{draganov2024hidden}.

In parallel, quantum machine learning (QML)~\cite{biamonte2017quantum} has emerged as a promising framework for learning expressive representation spaces via high-dimensional quantum feature mappings and entangled-state interactions. Quantum-inspired formulations, such as fidelity-based measures~\cite{liang2019quantum}, can yield nonlinear similarity signals by comparing parameterized quantum-state embeddings. Such state-overlap-based similarity is attractive for heterogeneous multimodal learning, where representations from different sensors may exhibit complex relationships that are difficult to capture using standard distance measures alone. Despite these advances, quantum-aware regularization in self-supervised predictive learning remains unexplored, especially in remote sensing and cross-modal representation learning.

As conceptually illustrated in Fig.~\ref{fig:teaser}, we propose HQ-JEPA, a Hybrid Quantum Joint-Embedding Predictive Architecture for cross-modal self-supervised learning in remote sensing. To the best of our knowledge, HQ-JEPA is the first framework to integrate quantum fidelity-based regularization into JEPA-style predictive representation learning for cross-modal remote sensing. HQ-JEPA combines JEPA-style masked latent prediction with cross-modal alignment, SIGReg-based~\cite{balestriero2511lejepa} distributional regularization, and a differentiable SWAP-test-based Fidelity Quantum Similarity (FQS) loss. Together, these objectives encourage context-aware prediction, modality-invariant semantic alignment, stable representation geometry, and fidelity-based similarity between representations. Our contributions are summarized as follows:
\begin{enumerate}
    \item We propose Hybrid Quantum Joint-Embedding Predictive Architecture (HQ-JEPA), a hybrid quantum-classical cross-modal self-supervised framework for remote sensing representation learning from paired Sentinel-1 and Sentinel-2 imagery.

    \item We extend JEPA-style masked latent prediction to a cross-modal setting and integrate SIGReg-based distributional regularization to improve semantic consistency, latent-space geometry, and training stability across heterogeneous sensing modalities.

    \item We introduce a Fidelity Quantum Similarity (FQS) loss, implemented using a differentiable SWAP-test quantum circuit, to provide a state-overlap-based similarity signal between latent representations.

    \item We evaluate HQ-JEPA on GeoBench downstream classification and segmentation tasks under linear probing and fine-tuning settings, demonstrating competitive and often superior performance over strong self-supervised and remote sensing foundation-model baselines.
\end{enumerate}

\section{Related Work}
\label{sec:related_work}

\subsection{Self-Supervised Visual Representation Learning}

Self-supervised learning has led to significant advances in learning visual representations from unlabeled data. Early contrastive methods such as SimCLR~\cite{chen2020simple} and MoCo~\cite{he2020momentum} learn representations by maximizing agreement between augmented views of the same image while contrasting them against negative samples. Although contrastive objectives have shown strong downstream performance, they generally require a large batch size or memory banks that hold negative samples for maximum performance~\cite{yokoo2021contrastive, chen2022we}. To reduce reliance on negative samples, non-contrastive approaches such as BYOL~\cite{grill2020bootstrap}, SimSiam~\cite{chen2021exploring}, and DINO variants~\cite{caron2021emerging,simeoni2025dinov3} use predictor asymmetry, stop-gradient operations, or momentum encoders to avoid representational collapse.

Masked image modeling (MIM) approaches, such as MAE~\cite{he2022masked} and BEiT~\cite{bao2021beit}, learn representations by reconstructing masked image patches. However, pixel-level reconstruction can encourage the model to focus on low-level details rather than high-level semantic structure. Predictive latent modeling approaches (e.g., I-JEPA~\cite{assran2023self}, VL-JEPA~\cite{chen2025vl}, and V-JEPA~\cite{mur2026v}) aim to directly predict representations of masked regions in feature space, rather than reconstructing pixels, leading to greater semantic abstraction and transferability of the learned representations.

Self-supervised learning has also been widely explored in remote sensing, where satellite imagery introduces challenges such as multispectral channels, multi-resolution observations, and sensor-dependent appearance variations. SatMAE~\cite{cong2022satmae} and ScaleMAE~\cite{reed2023scale} adapt masked autoencoding to multispectral and multiscale geospatial imagery, while recent foundation models such as AnySat~\cite{astruc2025anysat} aim to support multiple resolutions, scales, and modalities. Multi-modal approaches such as MMEarth~\cite{nedungadi2024mmearth} and DeCUR~\cite{wang2024decoupling} further exploit complementary information across Earth observation modalities. However, these methods primarily rely on reconstruction, contrastive, or representation decomposition objectives, whereas our work introduces quantum-fidelity-based regularization into JEPA-style cross-modal predictive representation learning.

\subsection{Representation Geometry and Distributional Regularization}
Recent research highlights the importance of representation geometry for robustness and generalization. Existing methods impose decorrelation or variance constraints using whitening~\cite{ermolov2021whitening}, covariance regularization, and redundancy reduction (e.g.,  Barlow Twins~\cite{zbontar2021barlow}, VICReg~\cite{bardes2021vicreg}) to reduce the likelihood of collapse and improve the conditioning of the embedding. The second-order statistics of embeddings are adjusted by these approaches, yet the global distributional structure is not explicitly enforced.

Several studies suggest that isotropic Gaussian embeddings would yield desirable optimization and transfer characteristics. The investigation of explicit distribution matching during self-supervised cross-modal learning has been limited. The SIGReg objective, proposed by Balestriero and LeCun~\cite{balestriero2511lejepa}, uses random projections to match the characteristic function of the embedding with that of an isotropic Gaussian to enforce Gaussian structure on the embeddings. In this work, we adapt SIGReg to a fused cross-modal latent space, encouraging a stable representation geometry for predictive alignment.

\subsection{Quantum and Quantum-Inspired Self-Supervised Learning}
Recently, hybrid quantum-classical models have been investigated for contrastive and self-supervised representation learning. Some works extend SimCLR-style frameworks by replacing classical projection heads with variational quantum circuits (VQCs)~\cite{cerezo2021variational}, where classical features are encoded in qubits via parameterized rotations and entanglement layers. These models are optimized using objectives such as NT-Xent with parameter-shift gradients~\cite{jaderberg2022quantum, ren2025quantum, padha2024qclr}. Other works have also investigated similar ideas in the domain of language modeling, where quantum circuits are now incorporated within the masked language model pre-training and fine-tuning pipeline~\cite{yao2025self}. Collectively, these studies demonstrate the potential of integrating quantum feature mappings with self-supervised objectives, particularly in low-data or noisy environments. Beyond standard benchmarks, hybrid quantum-classical models have also been explored in remote sensing for image classification~\cite{hossain2026qmc} and segmentation~\cite{hossain2026hq,hossain2026hqf}.

In addition to contrastive learning, fidelity-based quantum objectives have been proposed for unsupervised similarity learning. Methods such as QSEA~\cite{li2026qsea} and QUSL~\cite{2024arXiv240402028Y} rely on quantum state fidelity or Hilbert-space similarity to guide representation learning, while additional works investigate quantum-inspired architectures or search strategies for self-supervised circuits~\cite{konar2021qutrit, he2025self}. Though prior approaches have focused primarily on the contrastive pre-training of representations or implementing standalone quantum embedding models, the integration of fidelity-based quantum similarity into predictive latent architectures (e.g., JEPA-style masked modeling) has received limited attention. In this work, we incorporate a differentiable SWAP-test-based fidelity loss as a regularization signal within a cross-modal predictive self-supervised framework.

\section{Proposed Methodology}
\label{sec:methodology}

\subsection{Overview of the Architecture}
\begin{figure}[ht]
    \centering
    \includegraphics[width=\linewidth]{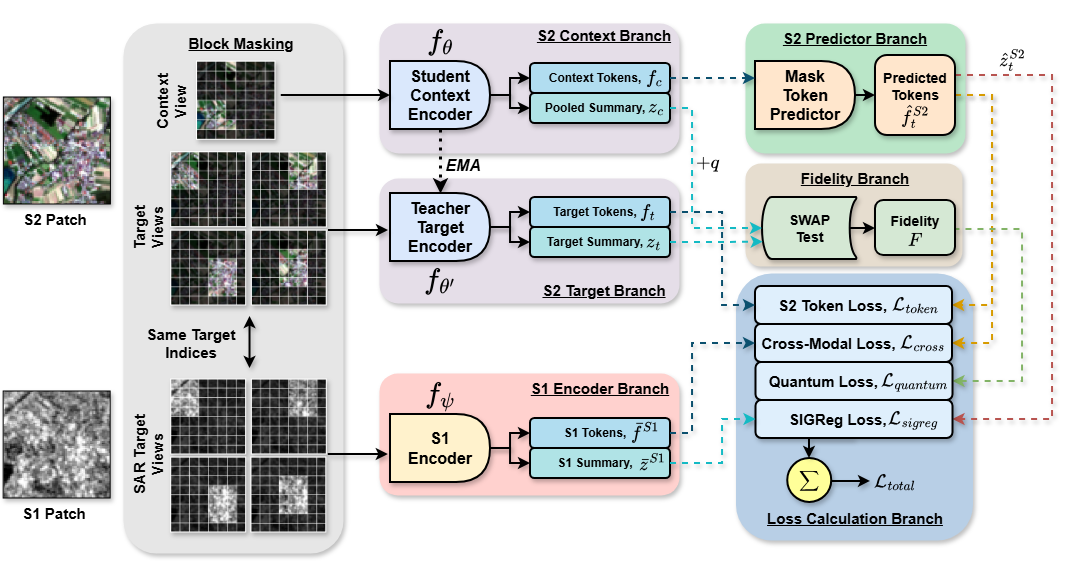}
    \caption{HQ-JEPA architecture for multi-modal remote sensing: S2 and S1 patches undergo block masking and are processed via student and teacher encoders for S2 (context and target branches) and S1 encoder branches. Outputs feed the Mask Token Predictor, Fidelity Branch, and combined loss branch, integrating token, cross-modal, quantum, and SIGReg.}
    \label{fig:hqjepa_arch}
\end{figure}

We present HQ-JEPA (Figure~\ref{fig:hqjepa_arch}), a hybrid cross-modal self-supervised framework that builds on predictive latent modeling by adding geometric and fidelity-based regularization objectives. For paired inputs $(x^{S1}, x^{S2})$, where Sentinel-2 (S2) is the primary modality and Sentinel-1 (S1) is the auxiliary modality, the model learns representations by predicting masked latent regions in feature space rather than reconstructing pixels.

\begin{figure}[ht]
    \centering
    \includegraphics[width=\linewidth]{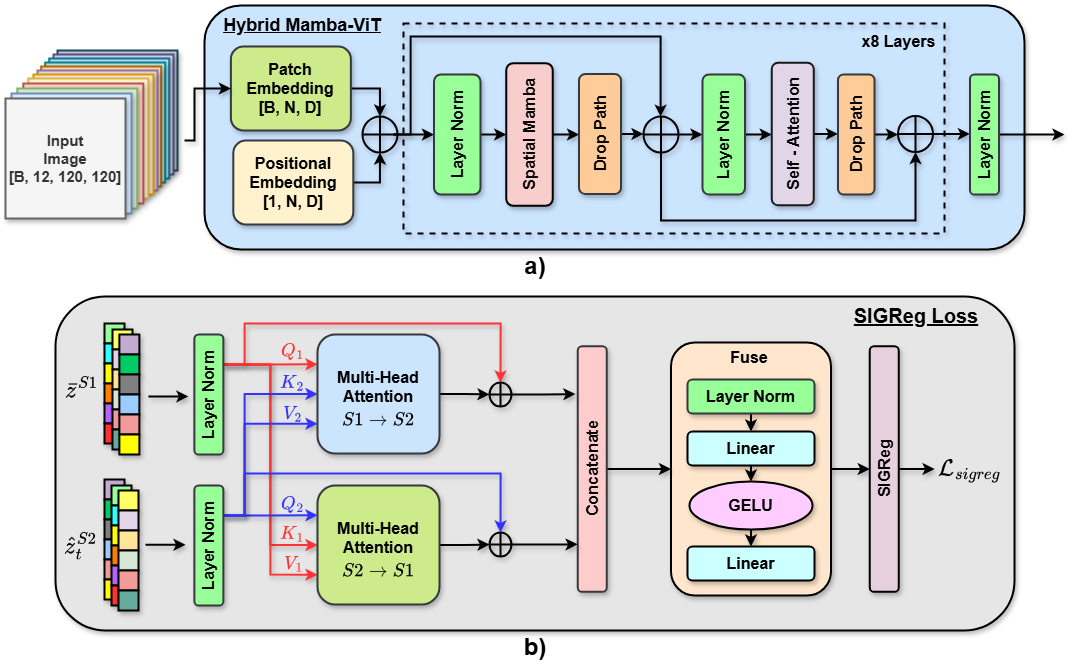}
    \caption{a) Hybrid Mamba-ViT encoder integrating patch/positional embeddings with stacked spatial Mamba and self-attention blocks for hierarchical feature learning; b) SIGReg module employing bidirectional cross-modal attention between S1 and S2 summaries, feature fusion, and a Gaussian regularization head for cross-modal representation alignment.}
    \label{fig:mamvit_sigreg}
\end{figure}

Block-based masking is applied to both $S2$ and $S1$ to sample visible context patches and multiple target regions. The visible patches from the $S2$ context view are processed by the Hybrid Mamba-ViT student context encoder, as shown in Fig.~\ref{fig:mamvit_sigreg}(a). The masked regions are encoded by a momentum-updated Teacher Target Encoder. A mask-token predictor predicts latent representations of masked regions using context tokens, and a loss is computed between the teacher target features and the predicted features.

For cross-modal alignment, the $S1$ modality is encoded separately and projected into the $S2$ embedding space. A token-level loss is calculated between the projected $S1$ features and the predicted $S2$ target tokens to preserve local semantic correspondence. Additionally, we introduce two additional regularization/alignment mechanisms: a differentiable SWAP-test-based fidelity loss and a cross-modal attention-based regularizer. The overall objective jointly promotes predictive, geometric, alignment, and fidelity-based terms.

\subsection{Block-Based Context and Target Masking}
\label{subsec:masking}
We adopt the block-based masking strategy from I-JEPA~\cite{assran2023self}, which separates visible context patches from multiple masked target patches. Given an input image $x_{S2}$, we first patchify it into non-overlapping patches. Let $\mathcal{P}$ denote the set of all patch indices. We sample 4 target blocks $\{\mathcal{T}_j\}_{j=1}^M \subset \mathcal{P}$, where each block represents a rectangular region which is contiguous in patch space. The block size and aspect ratio are randomly drawn from predefined ranges, and for proper supervision, each block must contain a specified minimum number of patches. The block location is sampled uniformly over the patch grid.


We then sample a context region $\mathcal{V} \subset \mathcal{P}$. It contains patches outside the target regions. By default, no overlap is allowed between context and target regions. This ensures:
\begin{equation}
    \mathcal{V} \cap \mathcal{T}_j = \emptyset, \quad \forall j.
\end{equation}
This encourages the model to predict latent representations for hidden patches solely from their neighboring visible content.

\begin{figure}[ht]
    \centering
    \includegraphics[width=\linewidth]{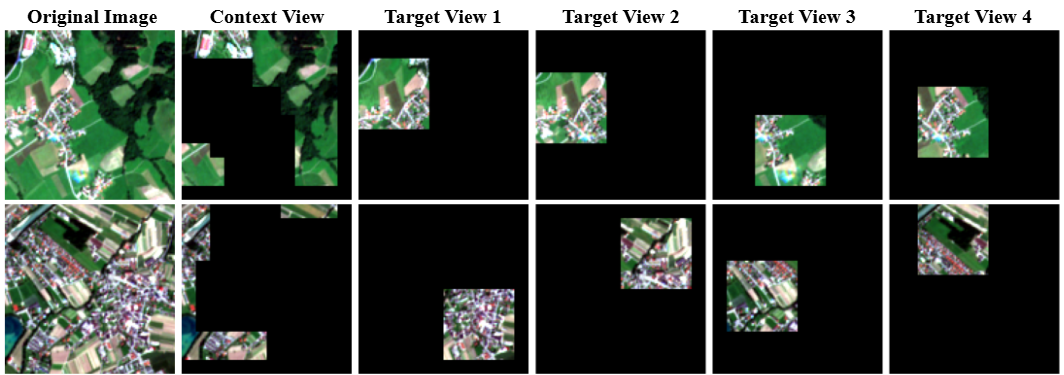}
    \caption{Illustration of the adopted I-JEPA masking strategy showing the original image, visible context region, and four target views used for predictive learning.}
    \label{fig:masking}
\end{figure}

Multiple target blocks are sampled per image during pretraining, while the context region is shared across them. Across both modalities, the same masks are applied consistently to preserve cross-modality alignment. In Figure~\ref{fig:masking}, we show one context view and four target views, which force the model to learn latent predictive relationships at various spatial locations. This type of block masking is different from simple random token masking. It enforces larger prediction regions and spatial coherence, which promotes higher-level semantic representation learning in these predictive architectures.

\subsection{Patch Token Predictive Alignment}
Given sets of masked patches, $\mathcal{V}$ (context) and $\{\mathcal{T}_j\}_{j=1}^M$ (targets), $x_{S2}$ is processed through a predictive student-teacher framework. The visible patches $x_{\mathcal{V}}^{S2}$ are passed through student context encoder $f_{\theta}$, producing contextual token representations:
\begin{equation}
    f_{c} = f_{\theta}(x_{\mathcal{V}}^{S2}).
\end{equation}
A pooled representation is obtained via average pooling:
\begin{equation}
    z_c = \text{AvgPool}(f_c).
\end{equation}

For each target block $\mathcal{T}_j$, the corresponding masked region is processed by a momentum-updated target encoder $f_{\theta'}$, where parameters $\theta'$ are updated as an exponential moving average of $\theta$:
\begin{equation}
    {f_t}^{(j)} = f_{\theta'}(x_{\mathcal{T}_j}^{S2}),\quad z_t^{(j)} = \text{AvgPool}(f_t^{(j)})
\end{equation}
The teacher target encoder uses stop-gradient to stabilize training. A mask-token predictor ($g_{\phi}$) takes context tokens with positional embedding corresponding to target patches and predicts their latent representations for each masked block:
\begin{equation}
    \hat{f_t}^{(j)} = g_{\phi}(f_c, \mathcal{V}, \mathcal{T}_j).
\end{equation}
The predictive objective minimizes a smooth regression loss between predicted and teacher target tokens:
\begin{equation}
    \mathcal{L}_{token} = \frac{1}{M} \sum_{j=1}^M \text{SmoothL1}(\hat{f}_t^{(j)}, f_t^{(j)}).
\end{equation}
This encourages the model to capture semantic relationships in latent space. The EMA teacher target encoder provides stable targets, while multiple target blocks promote diverse predictive learning across different spatial locations and semantic contexts.

\subsection{Cross-Modal Token Alignment}
For each target block $\mathcal{T}_j$, the corresponding region in $x_{S1}$ is processed by a modality-specific encoder $f_{\psi}$, yielding token-level representations:
\begin{equation}
    f_j^{S1} = f_{\psi}(x_{\mathcal{T}_j}^{S1}).
\end{equation}
These features are projected into embedding space of $S2$ student encoder via learnable projection head $h(\cdot)$:
\begin{equation}
    \bar{f}_j^{S1} = h(f_j^{S1}).
\end{equation}
Now, we impose a token-level regression loss between predicted S2 target tokens and projected S1 tokens:
\begin{equation}
    \mathcal{L}_{\text{cross}} = \frac{1}{M} \sum_{j=1}^{M} \text{SmoothL1} ( \bar{f}_j^{S1}, \hat{f}_{t,j}^{S2}).
\end{equation}

\subsection{Cross-Modal Gaussian Regularization}
Token-level alignment does not explicitly constrain the geometry of the learned embedding space. Anisotropic representations can degrade transfer stability and performance. To address this, we apply cross-modal Gaussian regularization (LeJEPA, SIGReg~\cite{balestriero2511lejepa}), which promotes isotropic latent geometry in a fused cross-modal space.

For each target block $\mathcal{T}_j$, we obtain pooled descriptors from both modalities:
\begin{equation}
    \bar{z}_{j}^{S1} = \text{Proj}(\bar{f}_j^{S1}), \quad \hat{z}_j^{S2} = \text{AvgPool}(\hat{f}_{t,j}^{S2}).
\end{equation}
These pooled descriptors are enhanced via bidirectional cross-attention, as shown in Figure~\ref{fig:mamvit_sigreg} b), allowing each modality to condition on the other. Then, both feature sets are fused via a learnable layer to produce a joint representation.


The distribution of these fused embeddings is regularized by matching their empirical characteristic function to that of an isotropic Gaussian. Given random projection directions $a \sim N(0, I)$ and scalar frequencies $t$, the characteristic function of a standard Gaussian is:
\begin{equation}
    \phi(t) = \text{exp} \left( - \frac{t^2}{2} \right).
\end{equation}
For projected embeddings, $x = a^{\top}z^{fuse}$, we compute:
\begin{equation}
    \mathcal{L}_{sigreg} = \mathbb{E}_{a,t} \left[(\mathbb{E}[\text{cos}(tx)] - \phi(t))^2 + (\mathbb{E}[\text{sin}(tx)])^2 \right].
\end{equation}

Using multiple random projection directions, the expectation and discretized frequencies are approximated. This regularization objective penalizes deviations from Gaussianity beyond second-order statistics, enforcing isotropic embeddings. In fused cross-modal space, SIGReg complements alignment and improves robustness of learned representations.

\subsection{Fidelity-Based Quantum Similarity (FQS) Loss}
Objectives like Euclidean distance and Cosine similarity measure first-order relationships between representations but do not capture higher-order correlations between feature dimensions. We introduce a differentiable fidelity-based quantum similarity objective enacted via a SWAP-test quantum circuit simulation. In quantum mechanics, the fidelity between two pure states $|\psi\rangle$ and $|\phi\rangle$ is defined as:
\begin{equation}
    F = | \langle \psi | \phi \rangle|^2,
\end{equation}
which quantifies state overlap in Hilbert space. The SWAP test provides a standard procedure for approximating fidelity using controlled-SWAP operations and an ancilla qubit.

In our framework, we encode representations into parameterized quantum states and compute fidelity as a non-linear overlap/similarity measure. Specifically, we extract context summary $z_c$ (student), target summary $z_t$ (teacher), and mask query embedding $q$ derived from target positional encodings. These feature vectors are projected into rotation parameters for quantum state preparation.

\begin{figure}[ht]
    \centering
    \includegraphics[width=\linewidth]{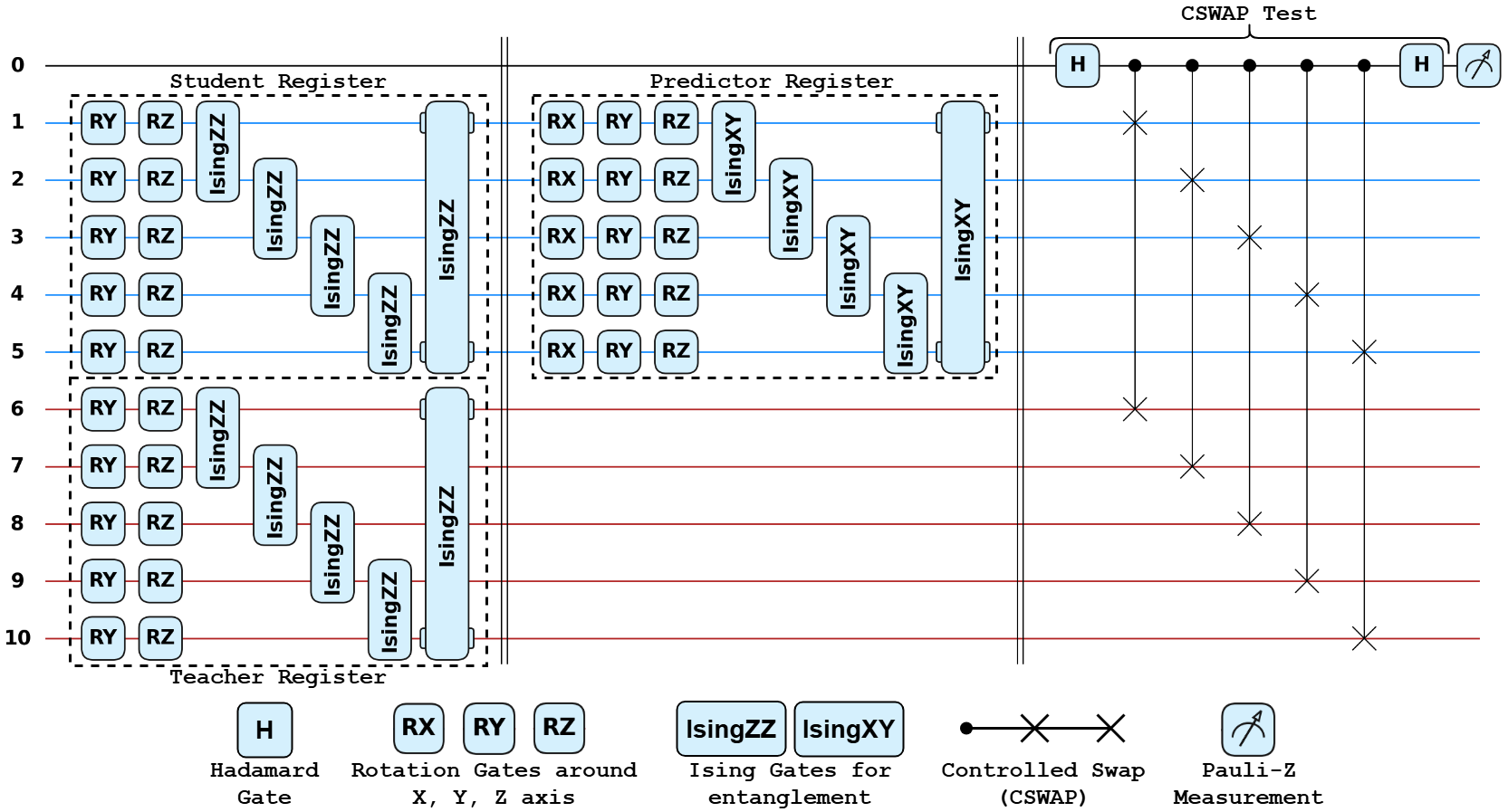}
    \caption{Quantum circuit for SWAP test between predicted student and teacher states: student qubits are modulated by predictor rotations and compared with teacher qubits using controlled-SWAP operations and Pauli-Z ancilla measurement.}
    \label{fig:swap_circ}
\end{figure}

We construct an 11-qubit circuit (Figure~\ref{fig:swap_circ}): 1 ancilla qubit, 5 student register qubits, and 5 teacher register qubits. The student and teacher summaries are encoded into layered parametrized rotations:
\begin{equation}
    U_{\text{enc}}(\theta) = \prod_{\ell=1}^{L} \left[ \bigotimes_{i=1}^{5} R_Y(\theta_{\ell,i}) R_Z(0.5\theta_{\ell,i}) \right] \cdot \text{IsingZZ}(\pi/4),
\end{equation}
where $\text{IsingZZ}$ gates produce ring entanglement within each register. The mask embedding $q$ modulates the student state through additional parametrized predictor layers:
\begin{equation}
    U_{\text{pred}}(q) = \prod_{\ell=1}^{L_p} \left[ \bigotimes_{i=1}^{5} R_X(q_{\ell,i,0}) R_Y(q_{\ell,i,1}) R_Z(q_{\ell,i,2}) \right] \cdot \text{IsingXY}(\pi/4).
\end{equation}

The SWAP test is then performed by initializing the ancilla qubit in superposition and applying controlled-SWAP gates pairwise between the student and teacher qubits. The ancilla is measured using the Pauli-Z expectation to estimate the fidelity as
\begin{equation}
    F = \frac{1 + \langle Z_{anc} \rangle}{2},
\end{equation}
and the quantum loss is defined as the corresponding infidelity:
\begin{equation}
    \mathcal{L}_{fqs} = 1 - F.
\end{equation}

The circuit is simulated using the adjoint differentiation method for efficient gradient computation. This design provides a structured higher-order similarity objective while maintaining scalability. The circuit mirrors the JEPA predictive structure: the context-derived student state and teacher target state are prepared in disjoint quantum registers with the same architecture. The predictor-modulated student state is then compared with the teacher target state using controlled-SWAP operations. This ancilla-mediated fidelity measurement plays a role analogous to the regression loss in JEPA, providing a Hilbert-space analogue of predictive latent matching.

\subsection{Unified Objective and Optimization}
HQ-JEPA is trained end-to-end by jointly optimizing predictive, geometric, cross-modal, and fidelity-based objectives. For a set of $M$ target blocks $\{\mathcal{T}_j\}_{j=1}^M$, the overall loss is:
\begin{equation}
    \mathcal{L}_{total} = \lambda_{t}\mathcal{L}_{token} + \lambda_{c}\mathcal{L}_{cross} + \lambda_{r}\mathcal{L}_{sigreg} + \lambda_{q}\mathcal{L}_{fqs}
\end{equation}
where, $\lambda_t=0.75$, $\lambda_c=0.5$, $\lambda_r=0.01$ and $\lambda_q=0.6$. To stabilize predictive training, the teacher target encoder $\theta'$ is updated using an exponential moving average (EMA):
\begin{equation}
    \theta' \leftarrow m\theta' + (1 - m)\theta,
\end{equation}
where $m \in [0, 1)$ is the momentum coefficient. All components-including projection heads, modality-specific encoders, and fusion modules-are jointly optimized via gradient descent.

\section{Results and Analysis}
\subsection{Experimental Setup}
HQ-JEPA was pretrained on the BigEarthNet dataset, which provides paired Sentinel-1 (S1) SAR and Sentinel-2 (S2) optical-multispectral imagery. S1 contains dual-polarization radar data, while S2 consists of 12 spectral bands, enabling cross-modal predictive learning from paired S1--S2 samples without manual annotations. Following Section~\ref{subsec:masking}, we apply block-based masking and optimize the full objective over paired S1--S2 inputs.

HQ-JEPA was trained for 400 epochs on 4 NVIDIA A100 80GB PCIe GPUs with a per-device batch size of 512 and gradient accumulation of 2, yielding an effective global batch size of 4096. We used a linear warm-up from $1\times10^{-5}$ to $3\times10^{-5}$ over 30 epochs, followed by cosine decay to $1\times10^{-7}$. Gradient clipping with a maximum norm of 1.0 was applied for stability, and the target encoder was updated using EMA with momentum 0.996. Following the I-JEPA masking protocol, we used an image size of $120 \times 120$, patch size 10, encoder mask scale $(0.85,1.0)$, predictor mask scale $(0.15,0.2)$, aspect ratio range $(0.75,1.5)$, and no overlap between context and target masks.

For downstream evaluation, we used the GeoBench~\cite{lacoste2023geo} benchmark, which provides standardized 12-band Sentinel-2 datasets for classification and segmentation. Classification tasks include m-BigEarthNet, m-EuroSAT, m-So2SAT, and m-Brick-Kiln, while segmentation tasks include m-SA-Crop and m-Cashew. For m-BigEarthNet evaluation, no downstream validation or test labels were used during self-supervised pretraining, and supervised evaluation followed the official GeoBench splits.

\subsection{Downstream Results}

Table~\ref{tab:main_results} compares HQ-JEPA with state-of-the-art remote sensing foundation models on GeoBench under Linear Probing (\twemoji{snowflake}) and Fine-Tuning (\twemoji{fire}) settings. The tasks cover diverse Earth observation applications, including multi-label classification, scene classification, binary classification, agricultural monitoring, and segmentation.

Under Linear Probing, HQ-JEPA demonstrates strong representation quality, achieving the best performance on m-bigearthnet (56.53), m-eurosat (82.59), and m-cashew (35.58), while remaining competitive on the remaining datasets. Notably, HQ-JEPA surpasses self-supervised approaches such as MAE, SatMAE, and AnySat on most tasks, indicating that the learned latent representations are transferable without updating encoder weights.

\begin{table}[ht]
\begin{center}
\footnotesize
\setlength{\tabcolsep}{2pt}
\begin{tabular}{|l|c|c|c|c|c|c|c|}
\hline

Method & Setting 
& \makecell{m-bigearthnet\\F1 ($\uparrow$)} 
& \makecell{m-eurosat\\Acc. ($\uparrow$)} 
& \makecell{m-so2sat\\Acc. ($\uparrow$)} 
& \makecell{m-brick-kiln\\Acc. ($\uparrow$)} 
& \makecell{m-SA-crop\\Jaccard ($\uparrow$)} 
& \makecell{m-cashew\\Jaccard ($\uparrow$)} \\

\hline

MAE \cite{he2022masked} & \twemoji{snowflake} 
& 55.41 & 78.00 & 44.42 & 88.89 & {\color{cyan}30.67} & 29.42 \\

SatMAE \cite{cong2022satmae} & \twemoji{snowflake} 
& 55.12 & 73.15 & {\color{cyan}46.04} & {\color{cyan}91.89} & 24.80 & 30.80 \\

MMEarth \cite{nedungadi2024mmearth} CNN-Atto & \twemoji{snowflake} 
& 43.30 & - & 43.80 & - & 22.20 & 24.20 \\

DeCUR \cite{wang2024decoupling} ViT-Small & \twemoji{snowflake} 
& - & - & - & - & 21.50 & 26.20 \\

AnySat \cite{astruc2025anysat} ViT-Base & \twemoji{snowflake} 
& - & - & - & - & 27.10 & 26.10 \\

\hline
\textbf{HQ-JEPA} & \twemoji{snowflake} 
& {\color{cyan}56.53} & {\color{cyan}82.59} & 45.96 & 91.47 & 30.48 & {\color{cyan}35.58} \\
\hline



MMEarth \cite{nedungadi2024mmearth} CNN-Atto & \twemoji{fire} 
& 67.10 & - & 54.60 & - & 38.20 & 79.80 \\

DeCUR \cite{wang2024decoupling} ViT-Small & \twemoji{fire} 
& {\color{orange}70.90} & {\color{orange}97.90} & 61.70 & {\color{orange}98.70} & - & - \\

AnySat \cite{astruc2025anysat} ViT-Base & \twemoji{fire} 
& 70.30 & 95.90 & 51.80 & 98.60 & - & - \\

DINO \cite{caron2021emerging} Resnet50 & \twemoji{fire}
& 67.37 & 96.79 & 54.39 & 98.43 & 37.26 & 83.09 \\

DOFA \cite{xiong2024neural} ViT-300M & \twemoji{fire}
& 68.58 & 96.18 & 61.26 & 98.22 & 35.90 & 81.07 \\

DeCUR \cite{wang2024decoupling} Resnet50 & \twemoji{fire}
& 67.54 & 97.61 & 56.68 & 98.27 & 34.49 & 84.15 \\




Prithvi-EO-2.0-100M \cite{szwarcman2025prithvi} & \twemoji{fire}
& 68.96 & 94.72 & 56.80 & 98.56 & 37.94 & 78.12 \\





Satlas-Swin-100M \cite{bastani2023satlaspretrain} & \twemoji{fire}
& 69.63 & 96.28 & 56.74 & 98.41 & 37.91 & 73.08 \\

ScaleMAE-ViT-300M \cite{reed2023scale} & \twemoji{fire}
& 62.80 & 91.55 & 48.33 & 98.05 & 25.75 & 73.60 \\

\hline

\textbf{HQ-JEPA} & \twemoji{fire} 
& 69.93 & 97.81 & {\color{orange}61.89} & 97.99 & {\color{orange}38.26} & {\color{orange}85.78} \\

\hline
\end{tabular}
\end{center}

\caption{Downstream GeoBench performance under {\color{cyan}Linear Probing} (\twemoji{snowflake}) and {\color{orange}Fine-Tuning} (\twemoji{fire}). Metrics are F1, accuracy, and Jaccard as indicated in the table header. A dash indicates that the result is not reported.}
\label{tab:main_results}
\end{table}

In the more challenging Fine-Tuning setting, HQ-JEPA achieves competitive or state-of-the-art performance across benchmarks. Specifically, it attains the highest scores on m-so2sat (61.89), m-SA-crop (38.26) and m-cashew (85.78), while achieving near-best performance on m-bigearthnet (69.93), m-eurosat (97.81) and m-brick-kiln (97.99). Despite training without task-specific architectural modifications, HQ-JEPA remains competitive with strong baselines including DeCUR, DINO, DOFA, Prithvi-EO-2.0, and Satlas, while achieving the best results on several downstream tasks.

Overall, these results demonstrate that HQ-JEPA learns robust, transferable, and semantically rich remote sensing representations, exhibiting strong generalization across diverse downstream tasks, evaluation protocols, and dataset domains.

\begin{figure}[ht]
    \centering
    \includegraphics[width=\linewidth]{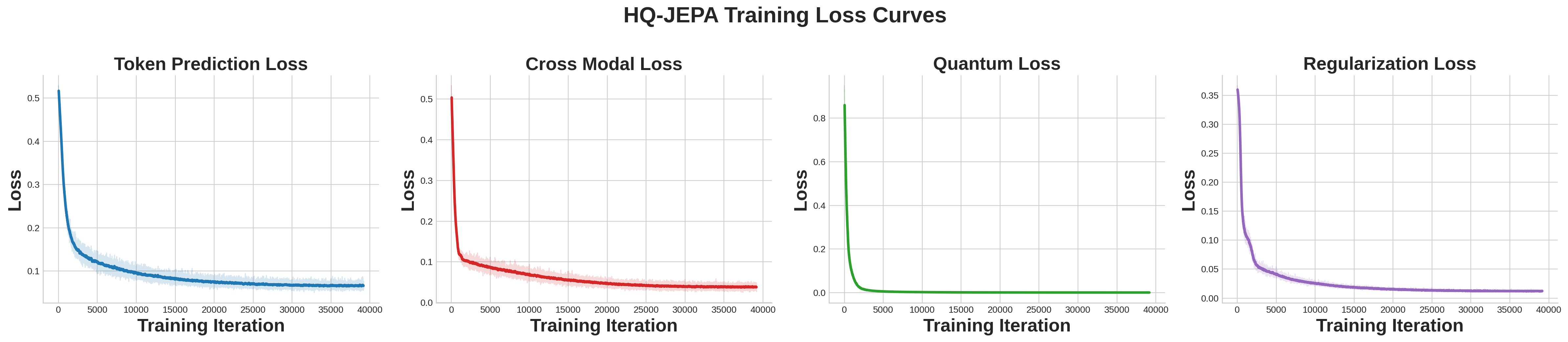}
    \caption{Training dynamics of HQ-JEPA over 400 epochs, showing the evolution of the token prediction, cross-modal, fidelity quantum similarity, and regularization losses.
    }
    \label{fig:loss_curves}
\end{figure}

Figure~\ref{fig:loss_curves} illustrates the training dynamics of the four HQ-JEPA objectives: token prediction, cross-modal alignment, Fidelity Quantum Similarity (FQS), and SIGReg regularization losses. The token prediction loss shows a rapid initial decrease followed by a smooth gradual decline, indicating effective context-aware latent prediction and steady refinement of semantic representations. Similarly, the cross-modal loss progressively decreases and stabilizes, suggesting successful alignment between heterogeneous modality representations in the shared latent space.

The FQS loss converges fastest, reaching a near-zero stable regime early and indicating rapid enforcement of latent overlap constraints. In contrast, the SIGReg loss decays gradually, reflecting progressive structuring of latent distributions. Overall, the stable convergence of all objectives suggests complementary interaction among predictive, alignment, regularization, and quantum losses for robust cross-modal representation learning.

\subsection{Ablation Studies}

Table \ref{tab:full_ablation} presents an ablation study evaluating the impact of each training objective in HQ-JEPA across GeoBench downstream tasks under both Linear Probing (LP) and Fine-Tuning (FT) settings. Starting from the base token prediction loss ($\mathcal{L}_{token}$), additional objectives are incrementally introduced to quantify their individual contributions to representation learning.

\begin{table*}[ht]
\begin{center}
\footnotesize
\setlength{\tabcolsep}{2pt}
\begin{tabular}{|c|c|c|c|cc|cc|cc|cc|cc|cc|}
\hline

$\mathcal{L}_{token}$ & $\mathcal{L}_{cross}$ & $\mathcal{L}_{sigreg}$ & $\mathcal{L}_{fqs}$ 
& \multicolumn{2}{c|}{\makecell{m-bigearthnet\\F1 ($\uparrow$)}} 
& \multicolumn{2}{c|}{\makecell{m-eurosat\\Acc. ($\uparrow$)}} 
& \multicolumn{2}{c|}{\makecell{m-so2sat\\Acc. ($\uparrow$)}} 
& \multicolumn{2}{c|}{\makecell{m-brick\\Acc. ($\uparrow$)}} 
& \multicolumn{2}{c|}{\makecell{m-SA-crop\\Jaccard ($\uparrow$)}} 
& \multicolumn{2}{c|}{\makecell{m-cashew\\Jaccard ($\uparrow$)}} \\
\cline{5-16}

& & &
& \twemoji{snowflake} & \twemoji{fire}
& \twemoji{snowflake} & \twemoji{fire}
& \twemoji{snowflake} & \twemoji{fire}
& \twemoji{snowflake} & \twemoji{fire}
& \twemoji{snowflake} & \twemoji{fire}
& \twemoji{snowflake} & \twemoji{fire} \\
\hline

\checkmark & & & 
& 51.24 & 65.11
& 76.85 & 94.82
& 41.13 & 56.34
& 89.12 & 96.71
& 27.42 & 34.15
& 27.08 & 79.93 \\

\checkmark & \checkmark & & 
& 52.61 & 66.42
& 78.14 & 95.47
& 42.38 & 57.29
& 89.83 & 97.03
& 28.36 & 35.12
& 28.01 & 81.06 \\

\checkmark & & \checkmark & 
& 53.47 & 67.31
& 79.26 & 96.11
& 43.57 & 58.44
& 90.38 & 97.32
& 29.14 & 36.01
& 29.05 & 82.37 \\

\checkmark & & & \checkmark
& 54.11 & 68.09
& 80.03 & 96.63
& 44.29 & 59.31
& 90.84 & 97.58
& 29.58 & 36.84
& 30.02 & 83.44 \\

\checkmark & \checkmark & \checkmark &
& 54.72 & 68.58
& 80.88 & 96.94
& 44.76 & 60.02
& 91.02 & 97.66
& 29.91 & 37.11
& 30.48 & 84.01 \\

\checkmark & & \checkmark & \checkmark
& 55.28 & 69.01
& 81.37 & 97.16
& 45.11 & 60.78
& 91.19 & 97.82
& 30.07 & 37.45
& 31.02 & 84.63 \\

\checkmark & \checkmark & & \checkmark
& 55.74 & 69.48
& 81.92 & 97.42
& 45.42 & 61.21
& 91.31 & 97.91
& 30.22 & 37.71
& 31.36 & 85.04 \\

\checkmark & \checkmark & \checkmark & \checkmark
& {\color{cyan}56.53} & {\color{orange}69.93}
& {\color{cyan}82.59} & {\color{orange}97.81}
& {\color{cyan}45.96} & {\color{orange}61.89}
& {\color{cyan}91.47} & {\color{orange}97.99}
& {\color{cyan}30.48} & {\color{orange}38.26}
& {\color{cyan}35.58} & {\color{orange}85.78} \\

\hline
\end{tabular}
\end{center}
\caption{Ablation study showing the contribution of each objective component across downstream tasks. For each dataset, values are reported as {\color{cyan}Linear Probing} (\twemoji{snowflake}) / {\color{orange}Fine-Tuning} (\twemoji{fire}).}
\label{tab:full_ablation}
\end{table*}

Using only $\mathcal{L}_{token}$ establishes a competitive baseline, achieving, for example, 51.24/65.11 on m-BEN (LP/FT), 76.85/94.82 on m-eurosat, and 27.08/79.93 on m-cashew. Incorporating the cross-view consistency loss $\mathcal{L}_{cross}$ yields consistent improvements across all tasks, increasing performance to 52.61/66.42 on m-bigearthnet and 78.14/95.47 on m-eurosat, indicating improved alignment between heterogeneous representations. Likewise, adding SIGReg regularization ($\mathcal{L}_{sigreg}$) further enhances downstream transferability by stabilizing the latent feature space.

A central finding of this study is the contribution of the proposed quantum fidelity similarity loss ($\mathcal{L}_{fqs}$). When added directly to the token objective, performance improves from 51.24 → 54.11 (LP) and 65.11 → 68.09 (FT) on m-bigearthnet, from 41.13 → 44.29 (LP) and 56.34 → 59.31 (FT) on m-so2sat, and from 27.08 → 30.02 (LP) and 79.93 → 83.44 (FT) on m-cashew. These gains are consistently larger than those obtained by using only cross-view alignment, demonstrating that the quantum objective contributes information beyond conventional predictive consistency mechanisms.

The quantum component becomes more evident in multi-objective settings. For example, comparing ($\mathcal{L}_{token} + \mathcal{L}_{cross} + \mathcal{L}_{sigreg}$) against configurations including quantum loss shows improvements across nearly all benchmarks. Notably, m-cashew FT improves from 84.01 to 85.78, m-so2sat FT increases from 60.02 to 61.89, and m-bigearthnet FT rises from 68.58 to 69.93 in the full model. These results suggest that quantum loss captures complementary structural information that strengthens feature expressiveness and generalization.

The complete objective, combining all four losses, achieves the best overall performance across all downstream tasks, reaching 56.53/69.93 on m-bigearthnet, 82.59/97.81 on m-eurosat, 45.96/61.89 on m-so2sat, 91.47/97.99 on m-brick, 30.48/38.26 on m-SA-crop, and 35.58/85.78 on m-cashew. Overall, the ablation validates that while each objective contributes positively to representation quality, the proposed quantum loss provides consistent and measurable gains, particularly on challenging transfer and segmentation benchmarks, making it a key component of HQ-JEPA.

\begin{table*}[ht]
\begin{center}
\footnotesize
\setlength{\tabcolsep}{2.5pt}
\begin{tabular}{|l|c|cc|cc|cc|cc|cc|cc|}
\hline
Setting & Value
& \multicolumn{2}{c|}{\makecell{m-bigearthnet\\F1 ($\uparrow$)}}
& \multicolumn{2}{c|}{\makecell{m-eurosat\\Acc. ($\uparrow$)}}
& \multicolumn{2}{c|}{\makecell{m-so2sat\\Acc. ($\uparrow$)}}
& \multicolumn{2}{c|}{\makecell{m-brick\\Acc. ($\uparrow$)}}
& \multicolumn{2}{c|}{\makecell{m-SA-crop\\Jaccard ($\uparrow$)}}
& \multicolumn{2}{c|}{\makecell{m-cashew\\Jaccard ($\uparrow$)}} \\
\cline{3-14}
 &
 & \twemoji{snowflake} & \twemoji{fire}
 & \twemoji{snowflake} & \twemoji{fire}
 & \twemoji{snowflake} & \twemoji{fire}
 & \twemoji{snowflake} & \twemoji{fire}
 & \twemoji{snowflake} & \twemoji{fire}
 & \twemoji{snowflake} & \twemoji{fire} \\
\hline

\multicolumn{14}{|c|}{\textbf{Encoder Depth}} \\
\hline

Depth & 4
& 53.84 & 66.92
& 79.31 & 95.86
& 43.12 & 58.74
& 89.42 & 96.84
& 28.11 & 35.06
& 29.14 & 82.43 \\

Depth & 6
& 55.26 & 68.71
& 81.24 & 97.05
& 44.58 & 60.61
& 90.71 & 97.51
& 29.62 & 36.84
& 30.73 & 84.31 \\

Depth & \textbf{8}
& {\color{cyan}56.53} & {\color{orange}69.93}
& {\color{cyan}82.59} & {\color{orange}97.81}
& {\color{cyan}45.96} & {\color{orange}61.89}
& {\color{cyan}91.47} & {\color{orange}97.99}
& {\color{cyan}30.48} & {\color{orange}38.26}
& {\color{cyan}35.58} & {\color{orange}85.78} \\

Depth & 10
& 55.81 & 69.12
& 81.88 & 97.28
& 45.11 & 61.03
& 91.03 & 97.64
& 29.97 & 37.24
& 31.18 & 84.86 \\
\hline

\multicolumn{14}{|c|}{\textbf{Number of Target Views}} \\
\hline

Views & 1
& 53.92 & 67.04
& 79.86 & 96.13
& 43.37 & 59.11
& 89.81 & 97.02
& 28.46 & 35.54
& 29.62 & 82.98 \\

Views & 2
& 54.98 & 68.36
& 80.97 & 96.92
& 44.29 & 60.33
& 90.58 & 97.46
& 29.54 & 36.63
& 30.48 & 84.05 \\

Views & 3
& 55.91 & 69.18
& 81.84 & 97.43
& 45.06 & 61.22
& 91.16 & 97.73
& 30.08 & 37.31
& 31.17 & 85.06 \\

Views & \textbf{4}
& {\color{cyan}56.53} & {\color{orange}69.93}
& {\color{cyan}82.59} & {\color{orange}97.81}
& {\color{cyan}45.96} & {\color{orange}61.89}
& {\color{cyan}91.47} & {\color{orange}97.99}
& {\color{cyan}30.48} & {\color{orange}38.26}
& {\color{cyan}35.58} & {\color{orange}85.78} \\
\hline

\multicolumn{14}{|c|}{\textbf{Target Scale}} \\
\hline

Scale & 0.125–0.2
& 54.71 & 68.08
& 80.34 & 96.57
& 43.96 & 59.87
& 90.24 & 97.18
& 29.13 & 36.02
& 30.11 & 83.61 \\

Scale & \textbf{0.15–0.2}
& {\color{cyan}56.53} & {\color{orange}69.93}
& {\color{cyan}82.59} & {\color{orange}97.81}
& {\color{cyan}45.96} & {\color{orange}61.89}
& {\color{cyan}91.47} & {\color{orange}97.99}
& {\color{cyan}30.48} & {\color{orange}38.26}
& {\color{cyan}35.58} & {\color{orange}85.78} \\

Scale & 0.2–0.25
& 55.64 & 69.04
& 81.42 & 97.19
& 44.87 & 60.92
& 90.88 & 97.53
& 29.92 & 37.21
& 31.06 & 84.69 \\

Scale & 0.2–0.3
& 54.92 & 68.31
& 80.76 & 96.74
& 44.15 & 60.08
& 90.43 & 97.31
& 29.41 & 36.52
& 30.54 & 83.98 \\
\hline

\multicolumn{14}{|c|}{\textbf{Context Scale}} \\
\hline

Scale & 0.4–1.0
& 54.46 & 67.84
& 80.11 & 96.38
& 43.82 & 59.74
& 90.12 & 97.06
& 28.94 & 35.91
& 29.94 & 83.42 \\

Scale & 0.65–1.0
& 55.51 & 68.96
& 81.34 & 97.17
& 44.78 & 60.88
& 90.92 & 97.61
& 29.87 & 37.06
& 30.97 & 84.56 \\

Scale & 0.75–1.0
& 56.11 & 69.51
& 82.01 & 97.54
& 45.42 & 61.36
& 91.21 & 97.82
& 30.19 & 37.54
& 31.44 & 85.21 \\

Scale & \textbf{0.85–1.0}
& {\color{cyan}56.53} & {\color{orange}69.93}
& {\color{cyan}82.59} & {\color{orange}97.81}
& {\color{cyan}45.96} & {\color{orange}61.89}
& {\color{cyan}91.47} & {\color{orange}97.99}
& {\color{cyan}30.48} & {\color{orange}38.26}
& {\color{cyan}35.58} & {\color{orange}85.78} \\
\hline
\end{tabular}
\end{center}
\caption{Hyperparameter ablation study on HQ-JEPA. For each dataset, values are reported as {\color{cyan}Linear Probing} (\twemoji{snowflake}) / {\color{orange}Fine-Tuning} (\twemoji{fire}); other parameters are fixed to default configuration.}
\label{tab:hyper_ablation}
\end{table*}

Table \ref{tab:hyper_ablation} presents a hyperparameter sensitivity analysis of HQ-JEPA under both Linear Probing (LP) and Fine-Tuning (FT) settings. We investigate the impact of four key design choices: encoder depth, number of target views, target scale, and context scale, while keeping all remaining parameters fixed to the default configuration.

We analyze the effect of encoder depth. Increasing depth from 4 to 8 layers consistently improves downstream performance, indicating that deeper encoders learn richer and more transferable representations. For instance, m-bigearthnet increases from 53.84/66.92 (LP/FT) at depth 4 to 56.53/69.93 at depth 8, while m-cashew improves from 29.14/82.43 to 35.58/85.78. However, increasing depth to 10 layers slightly degrades performance on several tasks, suggesting diminishing returns and possible over-parameterization. Consequently, Depth = 8 provides the best trade-off between model capacity and generalization.

Next, we study the effect of target views. Increasing the number of views progressively improves downstream performance, showing the benefit of predictive consistency across multiple latent targets. For example, m-so2sat improves from 43.37/59.11 with one target view to 45.96/61.89 with four views, while m-SA-crop increases from 28.46/35.54 to 30.48/38.26. This suggests that multi-view supervision strengthens representation learning by exposing the model to diverse predictive targets.

We also evaluate target scale, which controls the spatial extent of prediction regions. Very small targets (0.125--0.2) and larger targets (0.2--0.3) both reduce performance compared to the default setting. The intermediate scale of 0.15--0.2 achieves the strongest results, indicating an effective balance between local semantic detail and contextual difficulty.

Finally, we examine context scale, which determines the visible region used for prediction. Larger context regions consistently improve performance, with 0.85--1.0 yielding the best results. For instance, m-eurosat improves from 80.11/96.38 at 0.4--1.0 to 82.59/97.81 at 0.85--1.0, while m-brick increases from 90.12/97.06 to 91.47/97.99. This trend indicates that broader context supports more robust semantic representation learning.

Overall, the hyperparameter analysis shows that HQ-JEPA remains stable across configurations, while the default setting (Depth=8, Views=4, Target Scale=0.15--0.2, Context Scale=0.85--1.0) provides the strongest and most consistent GeoBench performance.
\subsection{Limitations}

HQ-JEPA still has several practical limitations. The fidelity module currently uses classical PennyLane simulation. This enables end-to-end differentiation but increases computational overhead compared to classical objectives, especially during long pretraining runs.

Because of this overhead, pretraining was done on approximately 237,871 image pairs from the BigEarthNet S1/S2 training split instead of the full archive. Expanding to the complete BigEarthNet dataset or other multimodal remote sensing collections may improve the learned representations and provide a better test of scalability.

The method also relies on manually chosen loss weights for the predictive, cross-modal, SIGReg, and quantum objectives. Although the current setup performs well on the evaluated GeoBench tasks, adjusting the loss weights or selecting more robust hyperparameters could make the framework easier to use across different datasets and combinations of modalities.
\section{Conclusion and Future Work}

We presented HQ-JEPA, a hybrid quantum-classical framework for cross-modal remote sensing representation learning in a self-supervised setting. The method extends JEPA-style masked latent prediction to paired Sentinel-1 and Sentinel-2 imagery and integrates four complementary objectives: token-level predictive learning, cross-modal alignment, SIGReg-based distributional regularization, and a differentiable SWAP-test-based Fidelity Quantum Similarity (FQS) loss. By operating directly in latent space, HQ-JEPA avoids pixel reconstruction and encourages semantically aligned, geometrically stable, and transferable representations across heterogeneous sensing modalities.

Experiments on GeoBench classification and segmentation tasks show that HQ-JEPA achieves strong performance under both linear probing and fine-tuning settings. The ablation study further confirms that each objective contributes to representation quality, with the quantum fidelity loss providing consistent gains when integrated with predictive and geometric regularization. These results suggest that quantum state-overlap-based regularization can serve as a useful complement to classical cross-modal self-supervised learning objectives.

Future work will focus on improving the scalability and applicability of HQ-JEPA. More efficient quantum simulation, hardware-aware circuit design, and adaptive loss balancing could reduce computational overhead and improve training stability. Scaling pretraining to the full BigEarthNet archive or larger multimodal Earth observation datasets may further enhance representation quality. Finally, extending HQ-JEPA to additional modalities, temporal data, and broader foundation-model pretraining remains a promising direction.

\bibliography{egbib}
\end{document}